\documentclass[runningheads]{llncs}
\usepackage[T1]{fontenc}
\usepackage{newtxtext}       % 
\usepackage[varvw]{newtxmath}       % selects Times Roman as basic font

\usepackage{booktabs}
\usepackage{enumitem}
\usepackage{multirow}
\usepackage{graphicx}
\usepackage{color}
\usepackage{hyperref}

\begin{document}
\title{Novelty Reviewer: A Large-Scale Dataset and Model for Reviewer-Aligned Novelty Assessment}
\titlerunning{Novelty Reviewer}
\author{Abeer Mostafa\inst{1}\orcidID{0000-0002-8971-4311} \and
Thi Huyen Nguyen\inst{2}\orcidID{0000-0001-8195-716\mathrm{X}} \and
Zahra Ahmadi\inst{1,3}\orcidID{0000-0003-1110-4756}}
\authorrunning{A. Mostafa et al.}
\institute{Peter L. Reichertz Institute for Medical Informatics, Hannover Medical School, Hannover, Germany\\
\email{\{Mansour.Abeer,Ahmadi.Zahra\}@mh-hannover.de}
\and
L3S Research Center, Hannover, Germany\\
\email{nguyen@l3s.de}
\and
Lower Saxony Center for Artificial Intelligence and Causal Methods in Medicine (CAIMed), Hannover, Germany}

\maketitle              % typeset the header of the contribution
\begin{abstract}
Assessing novelty in AI research is arguably the most consequential yet least reliable step in peer review. Reviewer judgments of novelty remain inconsistent and dependent on comparisons to prior work that are often incomplete.
We release a large-scale \textbf{Novelty Reviewer Dataset} built from $102{,}021$ official reviews of $37{,}899$ ICLR and NeurIPS submissions (2022--2025). For each submission, it provides the reviewers' novelty statements, a paper-level novelty assessment, and a novelty score on a four-point scale. 
On top of this dataset, we build \textbf{Novelty Reviewer}, a literature-aware framework that learns how reviewers judge novelty and grounds these judgments in semantic comparisons to existing research. We fine-tune a large language model on the dataset to capture reviewer-aligned novelty evaluation behavior. Given a manuscript, the framework extracts structured representations of its ideas, methods, and claims; retrieves semantically related papers; and constructs a knowledge graph that enables concept-level comparison to prior work. Based on the retrieved evidence, the model produces calibrated novelty scores alongside human-like explanatory assessments, reducing overestimation and improving consistency relative to existing approaches. Furthermore, we analyze different LLMs' abilities to assess originality and detect potential conceptual plagiarism. We publicly release the dataset, the model, and the code with documentation to support research on automated novelty assessment and peer review.\footnote{%
Project page: \url{https://m2l-lab.github.io/Novelty-Reviewer/}\newline
Dataset: \url{https://huggingface.co/datasets/M2L-Lab/Novelty-Reviewer-Dataset}\newline
Model: \url{https://huggingface.co/M2L-Lab/Llama-Novelty-Reviewer}\newline
Code and documentation: \url{https://github.com/m2l-lab/Novelty-Reviewer}}

\end{abstract}

\section{Introduction}

%“We are drowning in information but starved for knowledge.” -- John Naisbitt. This tension has never been more acute than in contemporary AI research. 
The rapid growth of AI research has fundamentally challenged our ability to assess originality at scale. 
In 2022, ICLR received 3,391 submissions~\cite{iclr2022factsheet}; by 2025, this number had surged to 11,603, representing a more than threefold increase in only three years~\cite{iclr2025factsheet}. 
As publication volume accelerates and reviewer workload intensifies in parallel, the assessment of research originality becomes increasingly difficult. 

On the other hand, reviewer guidelines for major venues explicitly require novelty judgments to be supported by concrete comparisons to prior work~\cite{iccv2025reviewerguidelines}. As a result, such assessments vary substantially across reviewers and are often constrained by incomplete coverage of the rapidly expanding literature~\cite{lin2023automated,sizo2025defining}. Empirical studies further confirm that reviewer judgments of originality can diverge considerably even for the same submission, exposing a critical inconsistency in the review process~\cite{teplitskiy2022novel}.
%Despite these limitations, novelty remains a central criterion in acceptance decisions, shaping research directions and determining what is recognized as a meaningful scientific contribution. %Assessing it demands that reviewers implicitly compare a submission against a vast, ever-expanding body of prior work, under severe time constraints and without standardized guidelines to anchor their judgment. 
Another concerning consequence is the silent acceptance of conceptually derivative work that repackages existing ideas through superficial reframing or terminological variation. Such work can evade conventional detection while presenting borrowed contributions as original. 

Despite these limitations, novelty remains a central criterion in acceptance decisions. Recent advances in LLMs have shown promise in automating aspects of peer review~\cite{dycke2023nlpeer,kuznetsov2024can,yu2024automated}. Existing automatic review-generation systems can generate fluent reviews; yet, their outputs tend to be overly positive and often lack deep conceptual critique as well as explicit reasoning about novelty~\cite{du2024llms,li2025unveiling,idahl2025openreviewer}. In particular, they do not generate solid originality assessments grounded in systematic comparisons with prior work. In parallel, prior approaches to novelty evaluation typically rely on embedding-based similarity~\cite{shibayama2021measuring,shahid2025literature} or classification signals~\cite{yan2025noveltyrank,zhao2025review}. While useful, these methods neither generate human-like analytical commentary nor provide interpretable comparison-based explanations that reflect how reviewers assess novelty in practice.

This paper addresses these challenges by introducing a reviewer-oriented framework for originality assessment, designed to generate human-like evaluative commentary that is explicitly grounded in systematic comparisons to relevant prior work.
We construct a large-scale dataset of peer review reports, from which we extract and aggregate reviewer discussions of novelty into paper-level originality assessments accompanied by normalized novelty scores. A critique-generation model is then trained to produce both categorized scores and free-text assessments aligned with human reviewing behavior. To ground these judgments, we incorporate a graph-based retrieval module that identifies conceptually related papers based on semantic similarity across ideas, methods, and claimed contributions. This retrieved context enables the model to refine and substantiate its generated assessments.
The resulting framework supports reviewers in producing evidence-based assessments by identifying original contributions, pinpointing overlaps with prior work, and articulating the basis for each judgment. It is important to note that the dataset and the framework are designed as a decision-support tool, assisting reviewers rather than replacing them. Our main contributions are as follows:
\begin{enumerate}[nosep] %,leftmargin=*
\item We release a dataset of $102{,}021$ peer reviews for $37{,}899$ ICLR and NeurIPS submissions (2022--2025). It pairs each submission with reviewer novelty statements, a paper-level novelty assessment, and a novelty score.
\item We develop a critique-generation model, trained on human reviewer judgments, that produces categorized novelty scores together with free-text justifications.
\item We introduce a graph-based retrieval module that identifies conceptually related prior work through semantic similarity across multiple aspects of the research; ideas, methods, and contributions.
\end{enumerate}

\section{Related Work}
A growing body of work has explored the automation of the peer-review process, as well as tools that assist reviewers, such as OpenReviewer~\cite{idahl2025openreviewer}. Domain-adapted LLMs, fine-tuned on large-scale peer review corpora, have demonstrated the ability to produce more structured and critical feedback compared to general-purpose LLMs~\cite{dycke2023nlpeer}. However, these approaches primarily focus on generating high-level review content, including overall ratings, strengths, and weaknesses, rather than providing fine-grained originality assessments grounded in specific prior work \cite{shin2025automatically}. 

Recent work has also investigated both the generation and evaluation of novelty using LLMs. On the generative side, Si et al.~\cite{sican} conduct a large-scale human evaluation comparing LLM-generated research ideas with those produced by expert NLP researchers. Their findings suggest that LLM-generated ideas are often perceived as more novel under blind evaluation, although concerns regarding feasibility remain. While this indicates that LLMs can approximate surface-level notions of originality, it does not address their ability to reliably assess or explain novelty, which is a critical and more practically useful task.

On the evaluation side, Lin et al.~\cite{lin2025evaluating} introduce SchNovel, a benchmark consisting of paper pairs across multiple domains, where the more recent paper is assumed to exhibit greater novelty. Their results show that retrieval-augmented methods improve novelty assessment by grounding model predictions in relevant literature. Complementary studies on LLM creativity \cite{zhao2025assessing,lullm,li2025automated} further demonstrate that originality remains the most challenging dimension to model, with improvements from multi-agent reasoning or structured prompting still limited by inherent subjectivity. 
%LLMs through Torrance-style tasks, role-play scenarios, and reference-based scoring

Another line of research approaches novelty from a computational perspective, leveraging embedding similarity, citation networks, and retrieval-augmented matching to estimate redundancy or atypicality relative to a corpus \cite{shibayama2021measuring,shahid2025literature,zhao2025review}. While effective for screening and ranking, these methods offer limited interpretability as similarity scores alone do not identify which idea or method overlaps, or why such overlap is meaningful. In this work, we address this limitation by modeling novelty assessment as a structured analytical task that integrates retrieval, alignment, and explicit comparison between a paper’s contributions and the most relevant prior work, thereby more closely reflecting how human reviewers reason about originality.

\section{Novelty Reviewer Dataset}
\subsection{Data Collection}
We collect submissions and their official reviews through the OpenReview API~\footnote{\url{https://openreview.net/}}. For each submission, we store its OpenReview identifier, full paper text, venue, and year. For each review, we store the full text and the numerical scores of the venue's review form. Review forms differ across venues and years, so we parse their scores into typed fields: overall rating and confidence, and sub-scores such as soundness, contribution, technical novelty, or originality. These sub-scores provide reviewer-assigned signals that complement our derived novelty labels.

\paragraph{\textbf{Anonymization}}
OpenReview data are public, yet we remove identifying information from all released text. Full paper texts are excluded, since camera-ready versions list authors, affiliations, and e-mail addresses. We match every text field against the author list of its submission and replace author names with a placeholder. Matching covers reversed and abbreviated forms. We also remove e-mail addresses, all links in abstracts, and links that contain an author's name. Reviewer identities are not part of the data, as OpenReview reviews are anonymous.

\paragraph{\textbf{Intended Use}}
The dataset supports three groups of tasks. First, novelty prediction and explanation, including predicting ECNA scores and generating reviewer-aligned novelty assessments. Second, related-work retrieval, where reviewer excerpts often name the prior work a submission is compared to, providing useful relevance signals. Finally, meta-research on peer review, such as studying reviewer agreement, score calibration, and how novelty judgments change across venues and years.

\subsection{Dataset Composition and Statistics} 
\label{sec:dataset}

\begin{figure}
    \centering
    \includegraphics[width=0.39\linewidth]{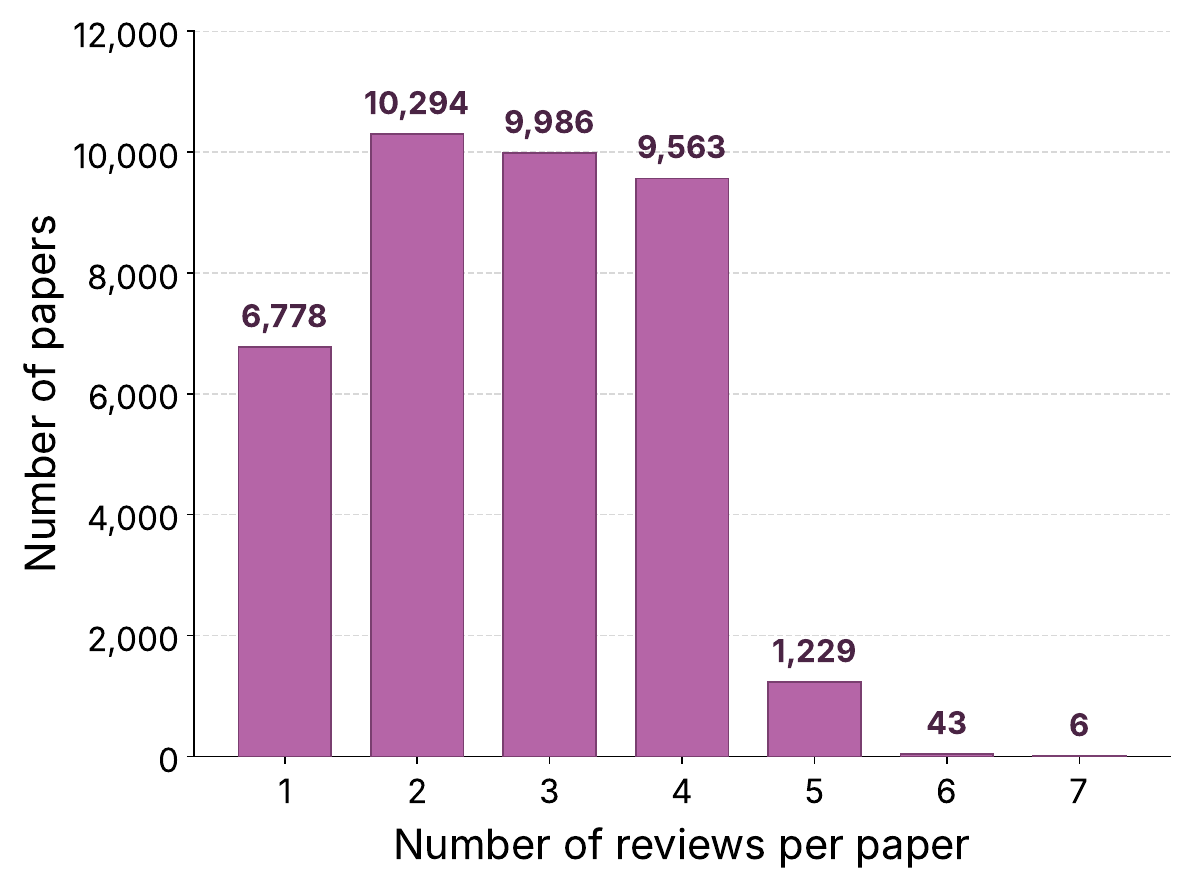}
    \caption{Distribution of the number of reviews per paper in the dataset.}
    \label{fig:reviews_per_paper}
\end{figure}

Our dataset links each submission to the novelty judgments of its reviewers. It covers $37{,}899$ submissions and $102{,}021$ official reviews from ICLR and NeurIPS, 2022--2025. It is one of the largest public resources of expert novelty judgments. Table~\ref{tab:data_dist} shows the composition by venue and year. The number of submissions grows from $4{,}963$ in 2022 to $15{,}934$ in 2025, reflecting the growth of both conferences. Each submission has $2.7$ reviews on average. 
\begin{table}[h]
\centering
\caption{Submissions and reviews per venue and year.}
\label{tab:data_dist}
\resizebox{0.5\linewidth}{!}{
\begin{tabular}{llrrrrr}
\toprule
\textbf{Venue} & & \textbf{2022} & \textbf{2023} & \textbf{2024} & \textbf{2025} & \textbf{Total} \\
\midrule
\multirow{2}{*}{ICLR}    & Papers  & 2,420 & 3,431 & 6,644  & 10,474 & 22,969 \\
                         & Reviews & 6,056 & 7,908 & 16,102 & 26,842 & 56,908 \\
\cmidrule{2-7}
\multirow{2}{*}{NeurIPS} & Papers  & 2,543 & 3,108 & 3,819  & 5,460  & 14,930 \\
                         & Reviews & 5,587 & 8,131 & 9,347  & 22,048 & 45,113 \\
\midrule
\multirow{2}{*}{Total}   & Papers  & 4,963  & 6,539  & 10,463 & 15,934 & \textbf{37,899} \\
                         & Reviews & 11,643 & 16,039 & 25,449 & 48,890 & \textbf{102,021} \\
\bottomrule
\end{tabular}
}
\end{table}
Figure~\ref{fig:reviews_per_paper} shows the number of reviews per paper. Most submissions received two to four reviews, in line with standard practice at both venues. NeurIPS 2025 shifts the distribution toward four reviews. Few papers received five or more reviews.
\subsection{Novelty-Related Text Extraction and Aggregation}
A two-stage procedure turns the reviews of each submission into a single, paper-level novelty assessment. We perform extraction step first, then aggregation.

\paragraph{\textbf{Extraction.}}
For each review, an instruction-tuned LLM (Llama-3.1-8B-Instruct) extracts all passages that bear on novelty. These include explicit statements, such as direct originality judgments or comparisons with prior work, and implicit signals, such as praise for a new formulation or criticism of an incremental contribution. The extractor copies passages verbatim, which avoids distortion through paraphrasing.

\paragraph{\textbf{Aggregation.}} The extracted passages of all reviewers of a submission are merged into one paper-level assessment. We distinguish three scenarios of reviewer agreement:
\begin{enumerate}[nosep,leftmargin=*]
    \item Consensus ($18{,}727$ papers): all reviewers agree on novelty. The model summarizes the shared view.
    \item Majority agreement ($12{,}477$ papers): reviewers disagree, but a clear majority supports one position. The model adopts the majority stance.
    \item Tie ($6{,}695$ papers): positive and negative views are evenly split. The model weighs the strength and specificity of the arguments on both sides, using reviewer confidence when available. Without decisive evidence, it assigns a marginal novelty score.
\end{enumerate}

Alongside the textual assessment, the model assigns a novelty score on an ordinal scale:
$-1$ (Not novel), the work is derivative and largely replicates existing approaches;
$0$ (Limited novelty), it makes minor variations of known methods;
$1$ (Moderately novel), it is original but overlaps notably with prior work or mainly extends existing ideas;
$2$ (Highly novel), it introduces fundamentally new ideas, methods, or problem formulations.
Scores reflect only the stance of the human reviewers, not external annotations or author claims. We call these labels \emph{Expert-Consensus Novelty Alignments} (ECNA). 

\subsection{Quality Validation}
We verify on $500$ sampled papers that aggregation preserves the reviewers' content. Each summary sentence is matched to its most similar review sentence using sentence embeddings~\cite{reimers-2019-sentence-bert}. The mean cosine similarity of these matches is $\mathbf{0.78}$. An entailment--contradiction (E--C) score from a natural language inference model reaches $\mathbf{0.79}$. Both results indicate close semantic alignment between the generated assessments and the source reviews.
We also run a human evaluation on a separate set of $500$ papers. An expert read the reviews of each paper and gave an overall novelty score based on reviewers opinions on the same scale used by the LLM. The expert's score matched the LLM-assigned score in $\mathbf{83\%}$ of cases. This also confirms that the aggregated novelty judgments in the dataset agree with how a human expert reads the same reviews.

\section{Novelty Reviewer}
\label{sec:method}
\begin{figure}
    \centering
    \includegraphics[width=0.99\linewidth]{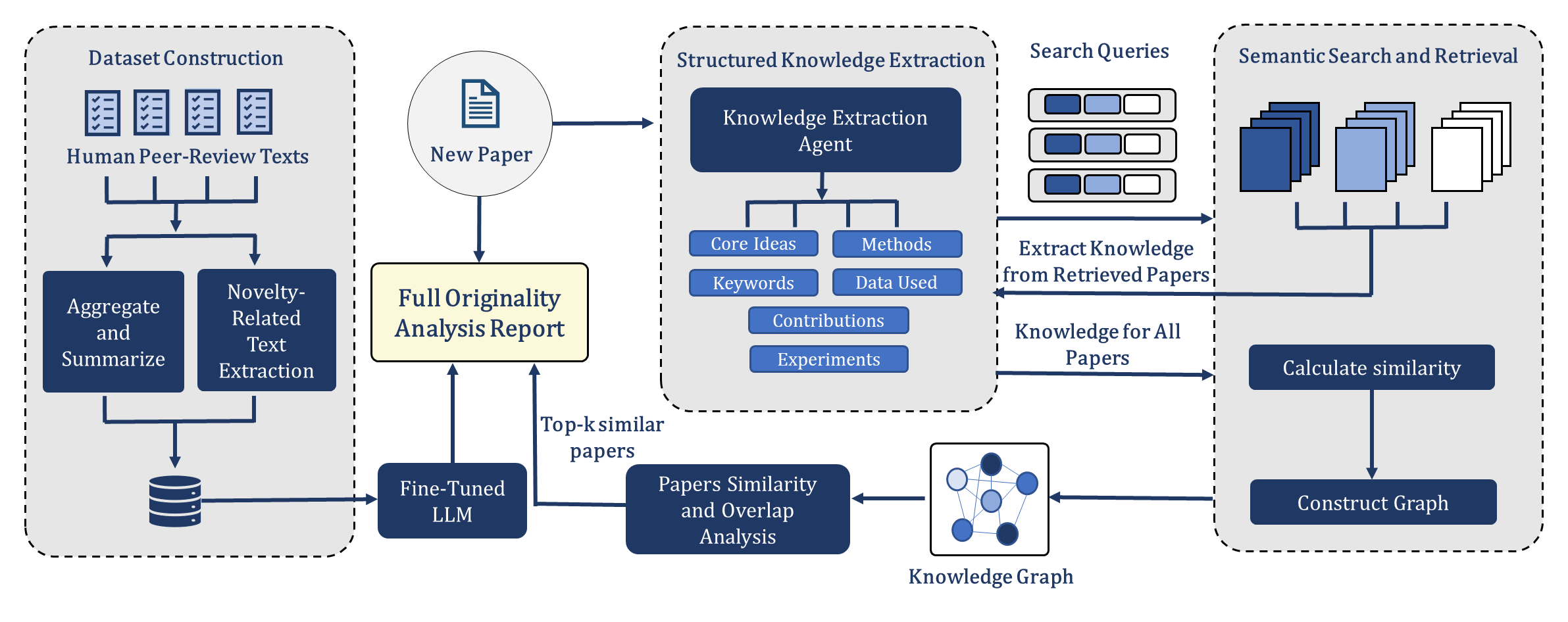}
    \caption{Full pipeline of Novelty Reviewer. Novelty-related passages from peer reviews are extracted and aggregated into paper-level originality assessments and normalized scores, forming a benchmark for fine-tuning. A knowledge extraction LLM structures each manuscript into ideas, methods, keywords, data, contributions, and experiments, to enable semantic retrieval of related papers. Pairwise field similarities establish links in a knowledge graph, from which the top-$k$ most relevant works are selected to ground the fine-tuned LLM’s final originality report.}
    \label{fig:method}
\end{figure}

Our framework for research originality analysis consists of two complementary stages. First, we learn patterns of novelty assessment from human expert reviews by training on the large-scale peer-review dataset. Second, we apply this learned signal to new manuscripts, combining retrieval-augmented reasoning with contrastive analysis against semantically related prior work. The overall pipeline is illustrated in Figure~\ref{fig:method}.

\subsection{Knowledge Extraction}

Given a target manuscript, we first transform its unstructured scientific prose into a structured semantic representation to enable principled comparison. Let $m$ denote the target manuscript. We employ a
knowledge-extraction LLM based on Llama-3.1-8B-Instruct,
which parses the full text of $m$ and produces a tuple:
$
K_{\text{ms}} = \langle C,\, M,\, R,\, D,\, E \rangle,
$
where $C$ denotes the set of \emph{core ideas}, $M$ \emph{methods}, $R$ \emph{contributions}, $D$ \emph{data sources}, and $E$ \emph{experimental components}. Each element in the tuple is a collection of semantically meaningful descriptors extracted directly from the manuscript. This representation serves two purposes. First, it makes the manuscript's intellectual content explicit and machine-comparable. Second, it provides the necessary query terms for the subsequent retrieval stage.

\subsection{Semantic Retrieval and Knowledge Graph Construction}

\paragraph{\textbf{Retrieval.}}
To contextualize the manuscript within the existing literature, we perform semantic retrieval using the Semantic Scholar API~\cite{fricke2018semantic}. Each component of $K_{\text{ms}}$ is issued as an independent query, retrieving up to five candidate papers per query. The union of retrieved documents forms the candidate set $P = \{p_1, \dots, p_n\}$. For each $p_i \in P$, the same knowledge-extraction LLM is applied to obtain a structured representation $K_i$ using an identical schema to $K_{\text{ms}}$. 

\begin{figure}[!h]
    \centering
    \includegraphics[width=0.56\linewidth]{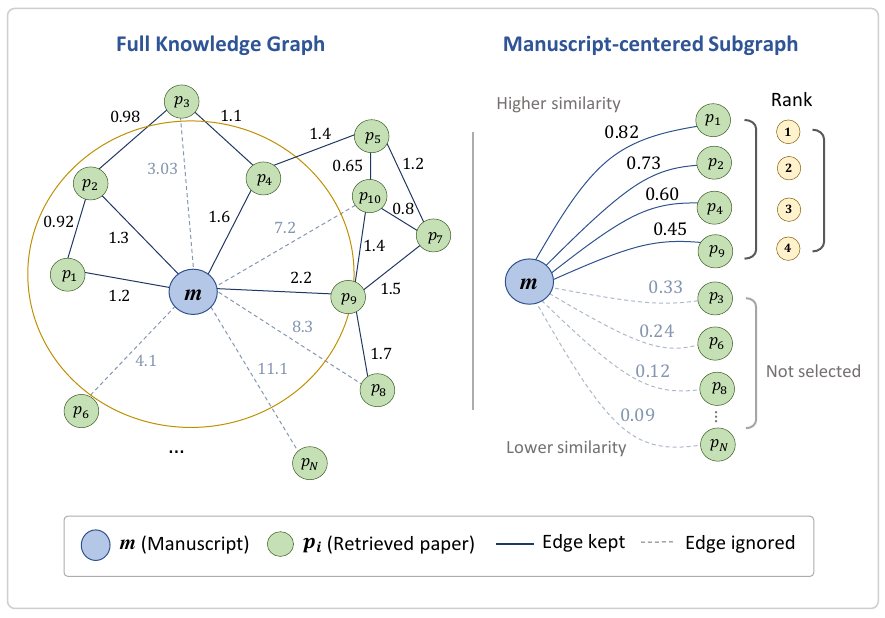}
    \caption{Knowledge graph construction and manuscript-centered ranking. Left: The full graph over the manuscript $m$ and all retrieved papers, with edges weighted by $w_{ij}^{\text{dist}}$. Right: Papers are ranked by weighted similarity to $m$; the top-$k$ (solid) form the comparative context $P_{\text{top}}$, while lower-ranked papers (dashed) are excluded.}
    \label{fig:kg}
\end{figure}

\paragraph{\textbf{Knowledge Graph.}}
We construct a knowledge graph $G = (V, E)$ (Figure~\ref{fig:kg}) where the node set $V = \{m\} \cup P$ contains the manuscript $m$ and all retrieved papers, and the edge set $E$ encodes pairwise semantic similarity. An edge $e_{ij}$ between nodes $(v_i, v_j)$ is included if their similarity exceeds a threshold $\tau$:
\begin{equation}
e_{ij} = \mathbf{1}\bigl[\text{Sim}(K_i, K_j) \geq \tau\bigr],    
\end{equation}
weighted by the similarity score $w_{ij} = \text{Sim}(K_i, K_j)$. 
To enable distance-based graph analysis, we define a complementary edge distance:
\begin{equation}
w_{ij}^{\text{dist}} = \frac{1}{\text{Sim}(K_i, K_j)}, \quad \forall\, e_{ij} \in E,
\end{equation}
where all retained edges satisfy $\text{Sim}(K_i, K_j) \geq \tau > 0$, and shorter distances reflect stronger conceptual proximity.
We select $\tau = 0.15$ based on a sensitivity sweep over $\tau \in \{0.05, 0.10, 0.15, 0.20, 0.25, 0.30\}$. Lower thresholds ($\tau \le 0.10$) admit an excessive number of papers into the knowledge graph, making construction computationally infeasible at scale; higher thresholds ($\tau \ge 0.25$) yield an overly sparse graph with too few retained neighbors to provide meaningful literature context. $\tau = 0.15$ was chosen as the best trade-off between graph density and retrieval relevance. 
Similarity is computed as the mean cosine similarity across all shared structured fields:
\begin{equation}
\text{Sim}(K_i, K_j) = \frac{1}{|F|} \sum_{f \in F} \cos\!\bigl(\phi(K_i^f),\, \phi(K_j^f)\bigr),
\end{equation}
where $F = \{C, M, R, D, E\}$ is the set of knowledge fields and $\phi(\cdot)$ denotes the sentence embedding of a field's concatenated descriptors.

The manuscript-centered subgraph $G_m \subseteq G$, comprising only edges incident to $m$, is used to rank retrieved papers with respect to $m$. The threshold $\tau$ defines a similarity radius around $m$: only papers within this radius are admitted as neighbors in $G_m$, ensuring that $P_{\text{top}}$ is drawn exclusively from papers with meaningful conceptual overlap. Papers are ranked by a composite score combining $70\%$ direct similarity to $m$ and $30\%$ within-graph centrality computed over inter-paper edges in $G$, prioritising papers that are both semantically close to the manuscript and central within the retrieved literature. The top-$k$ papers by this score form the comparative context $P_{\text{top}}$.

\subsection{Fine-Tuning for Novelty Estimation}

To model expert reasoning about originality, we fine-tune Llama-3.1-8B-Instruct on our large-scale novelty benchmark. Unlike zero-shot prompting approaches, where the model's prior is shaped by generic pre-training rather than the norms of scientific peer review, fine-tuning aligns the model with its judgments with the vocabulary, reasoning, and calibration of expert reviewers. 
Each training instance consists of the full manuscript text as input and a structured target comprising the aggregated novelty assessment, the extracted novelty-relevant reviewer statements, and a normalized novelty score in $\{-1, 0, 1, 2\}$. 

\subsection{Originality Report Generation}

At inference time, the manuscript text, together with the retrieved knowledge representations of $P_{\text{top}}$, is passed as input to the fine-tuned model. Conditioning on both the manuscript content and the explicitly retrieved comparative context allows the model to produce novelty judgments that are grounded in concrete prior-work comparisons. The model outputs a calibrated novelty score and natural-language justifications that identify and describe the specific conceptual contrasts between the manuscript and the most relevant retrieved papers.
The final originality report integrates all analytical components of the pipeline into a unified document comprising. First, a novelty score with a short evidence-based justification summarizing the manuscript's originality relative to the retrieved literature. Second, an extracted knowledge summary of the manuscript's core ideas, methods, and keywords. Third, a ranked list of the most similar prior works, each presented with their similarity score, publication year, citation count, and a summary of their core ideas and methods to enable direct comparison with the manuscript.

\section{Experiments}
\label{sec:experiments}
\subsection{Experimental Setup}
\paragraph{\textbf{Training Procedure.}}
The Novelty Reviewer model is obtained by fine-tuning meta-llama/Llama-3.1-8B-Instruct on our novelty reviewer dataset. Training is performed on the full tokenized dataset, which contains aggregated peer-review supervision instances derived from our dataset construction pipeline. 
The training corpus consists exclusively of peer-review–derived supervision, where each example includes the manuscript text, extracted novelty-related review statements, and normalized novelty labels. All evaluations are conducted on a held-out test set.
To support long-context modeling, we set the maximum sequence length to 32,120 tokens, enabling the model to process full manuscripts and aggregated review contexts without truncation.
Training is conducted in mixed precision with automatic bfloat16 support and TensorFloat-32 acceleration. Weight decay is disabled, following standard practice for instruction fine-tuning of large language models.
Training is executed using multi-GPU distributed execution (32 A100 GPUs with 80 GB memory each) with DeepSpeed ZeRO Stage-3 optimization. Additional efficiency improvements are provided via Liger kernel integrations, including optimized rotary embeddings, RMS normalization, GLU activations, and fused linear cross-entropy operations. The model follows the Llama-3.1-8B-Instruct chat template. Checkpoints are saved twice per epoch, and only model weights are retained to reduce storage overhead.

\paragraph{\textbf{Baselines.}} We compare against a diverse set of competitive baselines. General-purpose LLMs include GPT-OSS-20B~\cite{agarwal2025gpt}, Llama-3.1-8B-Instruct~\cite{grattafiori2024llama}, Mistral-7B-Instruct-v0.1 ~\cite{jiang2023mistral}, and Qwen2.5-14B-Instruct-1M~\cite{yang2025qwen25}, collectively covering a range of scales and architectures. Domain-adapted baselines include SciLlama~\cite{senthilkumar2025scillama32}, a science-focused language model; Paper Reviewer~\cite{weathon2025paperreviewer}, a Qwen3-8B model fine-tuned for review generation; and OpenReviewer~\cite{idahl2025openreviewer}.

\paragraph{\textbf{Evaluation.}} We train on the training split ($25{,}951$ papers) and evaluate on a held out $500$-paper evaluation set, which covers all novelty categories. We evaluate both \emph{discrete novelty score prediction}, measured as agreement with reviewer consensus, and \emph{free-text originality assessment}, measured as the semantic fidelity and consistency of generated justifications with expert consensus. We test \textbf{Novelty Reviewer} in two configurations: \textbf{w/o retrieval}, which isolates the effect of fine-tuning, and \textbf{full framework}, which adds retrieval augmentation.
For discrete novelty score prediction, we report Accuracy, Precision, Recall, and F1 score across the four novelty categories. Given the naturally skewed label distribution, where most peer-reviewed papers fall into moderate novelty levels, Precision and F1 are particularly informative metrics. For free-text evaluation, we adopt Entailment--Contradiction Natural Language Inference (NLI) score, BERTScore~\cite{zhang2019bertscore},
which measures token-level semantic overlap with the reference ECNA justification, AlignScore~\cite{zha2023alignscore}, which
measures factual consistency between the generated text and the source material, and an LLM-as-a-Judge score using Llama-3.1-8B-Instruct as the evaluator, which assesses the correctness, coverage, and consistency between the generated answer and the reference and gives a score from 1 to 10.
\paragraph{Entailment--Contradiction (E--C) NLI Evaluation Metric:}
Each model-generated output is treated as the \emph{premise} and the corresponding reference summary as the \emph{hypothesis}. A pretrained Roberta-based NLI classifier produces probabilities over \emph{entailment}, neutral, and \emph{contradiction} for each pair $(p_i, h_i)$: 
$\mathbf{q}_i = [q_i^{\text{contra}}, q_i^{\text{neutral}}, q_i^{\text{entail}}]$.
We aggregate these scores across the evaluation set and compute the E--C score as
\begin{equation}
\mathrm{E\text{-}C} = \frac{1}{2}\left( \frac{1}{N}\sum_{i=1}^{N}
\left(q_i^{\text{entail}} - q_i^{\text{contra}}\right) + 1 \right),
\end{equation}
where $N$ denotes the number of evaluated samples. The affine transformation normalizes the metric to $[0,1]$, with higher values indicating stronger entailment and lower contradiction between the model output and the reference. Intuitively, this metric rewards outputs that are semantically supported by the reference novelty summary while penalizing contradictory statements.

\subsection{Results}
\begin{figure}
    \centering
    \includegraphics[width=0.55\linewidth]{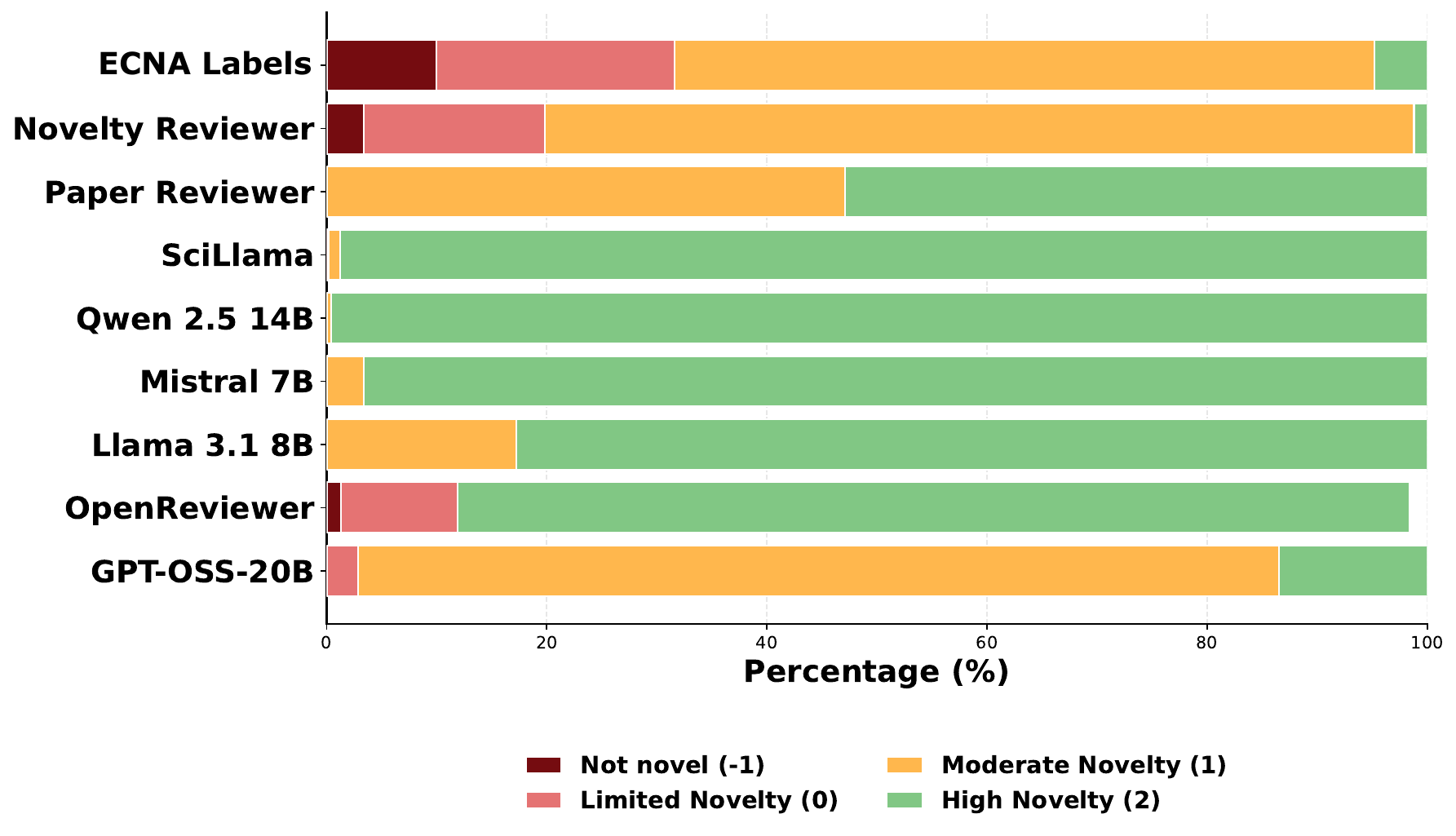}
    \caption{Predicted novelty score distribution across models.}
    \label{fig:perc}
    \vspace{-11pt}
\end{figure}

Table~\ref{tab:res} presents the full quantitative comparison across all models and metrics. The proposed Novelty Reviewer achieves the strongest performance across all evaluation dimensions in both configurations, with the full retrieval-augmented variant yielding the best overall results.
We note that absolute F1 scores remain moderate because of the difficulty of the underlying 4-class task under a label distribution skewed toward moderate novelty. Despite this, Novelty Reviewer achieves a good relative F1 improvement over every baseline, including frontier general-purpose LLMs, and is the only model whose predicted score distribution closely tracks the ECNA distribution (Figure~\ref{fig:perc}), which we view as the more informative comparison given the inherent class imbalance.
Beyond scalar prediction, qualitative evaluation via E--C NLI, BertScore, AlignScore, and LLM-as-a-Judge all confirm that the Novelty Reviewer generates justifications that are more consistently grounded in reference rationales than all baselines. 

\paragraph{\textbf{Generalization to Unseen Venues.}}
To assess whether the proposed framework generalizes to other venues in a different domain, we conduct an evaluation experiment on submissions from the Medical Imaging with Deep Learning (MIDL) 2026 conference~\cite{midl2026}. MIDL represents a substantively different venue from the NeurIPS and ICLR data used during training as it targets a specialized interdisciplinary community at the intersection of computer vision, deep learning, and clinical medicine, with distinct domain vocabulary and novelty criteria.
We collect all 255 publicly available submissions from MIDL 2026 via the OpenReview API and extract all associated official reviews following the same procedure. Results are reported in the MIDL 2026 section of Table~\ref{tab:res}. Across all evaluation dimensions, the Novelty Reviewer consistently outperforms all baselines, despite never having been trained on this venue's data.

\begin{table}[ht]
\centering
\caption{Performance comparison of different models on
novelty score prediction and free-text evaluation.}
\label{tab:res}
\resizebox{0.95\linewidth}{!}{
\begin{tabular}{llcccccccc}
\toprule
\textbf{Data} &
\textbf{Model} &
\textbf{Accuracy} &
\textbf{Precision} &
\textbf{Recall} &
\textbf{F1 Score} &
\textbf{(E$-$C) NLI} &
\textbf{LLM Judge} &
\textbf{BERTScore} &
\textbf{AlignScore} \\
\midrule
\multirow{9}{*}{\rotatebox{90}{\textbf{Novelty Benchmark}}} & GPT-OSS-20B
& 0.53 & 0.2053 & 0.2950 & 0.2268 & 0.6659 & 7.033 & 0.72 & 0.79 \\

&Llama-3.1-8B-Instruct
& 0.15 & 0.1621 & 0.2797 & 0.0894 & 0.4832 & 5.492 & 0.67 & 0.65 \\

&Mistral-7B-Instruct-v0.1
& 0.07 & 0.1537 & 0.2574 & 0.0389 & 0.5529 & 3.566 & 0.55 & 0.58 \\

&Qwen2.5-14B-Instruct-1M
& 0.05 & 0.2620 & 0.2516 & 0.0261 & 0.4498 & 6.992 & 0.48 & 0.52 \\

&SciLlama
& 0.06 & 0.2121 & 0.2531 & 0.0294 & 0.5436 & 5.614 & 0.63 & 0.60 \\

&Paper Reviewer
& 0.33 & 0.1720 & 0.2893 & 0.1613 & 0.6816 & 5.871 & 0.71 & 0.77 \\

&OpenReviewer
& 0.08 & 0.1991 & 0.2026 & 0.0850 & 0.5139 & 6.186 & 0.62 & 0.67 \\

&\textbf{Novelty Reviewer (w/o retrieval)} & \textbf{0.60} & \textbf{0.3125} & \textbf{0.3066} & \textbf{0.3012} & \textbf{0.7421} & \textbf{7.523} &\textbf{ 0.77 }&\textbf{ 0.81} \\
&\textbf{Novelty Reviewer (full framework)} & \textbf{0.62} & \textbf{0.3777} & \textbf{0.3187} & \textbf{0.3231} & \textbf{0.7603} & \textbf{7.824} & \textbf{0.78} & \textbf{0.84} \\
\midrule
\multirow{9}{*}{\rotatebox{90}{\textbf{MIDL 2026}}} & GPT-OSS-20B
& 0.61 & 0.2384 & 0.2417 & 0.2479 & 0.6913 & 7.184 & 0.73 & 0.79 \\

&Llama-3.1-8B-Instruct
& 0.22 & 0.1815 & 0.2268 & 0.1176 & 0.5214 & 5.883 & 0.69 & 0.68 \\

&Mistral-7B-Instruct-v0.1
& 0.11 & 0.1692 & 0.2143 & 0.0621 & 0.5741 & 4.021 & 0.57 & 0.55 \\

&Qwen2.5-14B-Instruct-1M
& 0.09 & 0.2718 & 0.2331 & 0.0413 & 0.4687 & 7.102 & 0.49 & 0.46 \\

&SciLlama
& 0.10 & 0.2264 & 0.2299 & 0.0482 & 0.5568 & 5.947 & 0.64 & 0.61 \\

&Paper Reviewer
& 0.41 & 0.2017 & 0.2476 & 0.1891 & 0.7034 & 6.214 & 0.72 & 0.76 \\

&OpenReviewer
& 0.14 & 0.2146 & 0.1984 & 0.0938 & 0.5362 & 6.447 & 0.63 & 0.68 \\

& \textbf{Novelty Reviewer (w/o retrieval)} & \textbf{0.73} & \textbf{0.2921} & \textbf{0.2552} & \textbf{0.2801} & \textbf{0.7422} & \textbf{7.613} & \textbf{0.78} & \textbf{0.80} \\
& \textbf{Novelty Reviewer (full framework)} & \textbf{0.76} & \textbf{0.3033} & \textbf{0.2721} & \textbf{0.2902} & \textbf{0.7512} & \textbf{7.664} &\textbf{ 0.79} &\textbf{ 0.81} \\
\bottomrule
\end{tabular}
}
\end{table}

\paragraph{\textbf{Distributional Analysis.}}
Figure~\ref{fig:perc} compares the distribution of predicted novelty scores across all models with the human-derived expert-consensus novelty alignments. All baseline models exhibit an obvious bias toward assigning moderate or high novelty scores ($1$ or $2$). This bias indicates a frequent overestimation of originality by current LLMs.
In contrast, the Novelty Reviewer demonstrates greater sensitivity to low-novelty cases ($-1$ and $0$), producing a distribution that closely aligns with the empirical class proportions observed in peer-review annotations.

\subsection{Use Case: Idea-Level Plagiarism Detection}
To test whether models reason about conceptual content or only match surface wording, we randomly select 10 papers published in 2023. We paraphrase each one so that it keeps the original ideas, methods, and claims but uses different vocabulary and sentence structure. Each model is asked to assess the novelty of the paraphrased paper and to identify any overlapping prior work.
Novelty Reviewer identifies the source paper in 9 of 10 cases and assigns a negative novelty score in all 9. The strongest baselines, GPT-OSS-20B and Qwen2.5-14B-Instruct-1M, each recognize only 4 sources and penalize 3 and 2, respectively. OpenReviewer detects and penalizes 2 cases. Llama-3.1-8B-Instruct, Paper Reviewer, and SciLlama detect 1 case each, and SciLlama does not penalize the one it detects. Mistral-7B-Instruct-v0.1 detects none. 

\section{Limitations}
While our framework provides a structured and human-aligned approach to assessing research novelty, it is subject to some limitations that should be acknowledged. The retrieval pipeline currently depends on the coverage and quality of external scholarly databases (Semantic Scholar). Additionally, our data comes from NeurIPS and ICLR. This makes the model more suitable for AI and ML research themes.
The dataset inherits the subjectivity of peer review as ECNA labels encode reviewer opinion, not ground-truth novelty. Finally, the dataset covers only English-language machine learning venues.

\section{Ethical Considerations}
This work addresses the sensitive task of assessing research novelty and originality, which has potential implications for scientific evaluation and decision-making. Our system is designed to support, rather than replace, human judgment. Rather than acting as an automated final decision-maker, the framework serves as a complementary analytical tool that promotes consistency and transparency in novelty evaluation.
The dataset is built exclusively from publicly available OpenReview data. We release no full paper texts and remove author names, e-mail addresses, and identifying links from all released fields. Reviewers are anonymous on OpenReview, and the data contain no reviewer identities. The system evaluates novelty against accessible prior work and learned reviewer patterns; no author or affiliation information is used for fine-tuning or testing.

\section{Conclusion}
Novelty assessment remains among the most consequential and least reliable components of peer review. In this work, we introduce the Novelty Reviewer Dataset, a large-scale resource that pairs reviewers’ novelty statements with paper-level assessments and scores, offering a structured view of how the research community judges originality. Building on this resource, we develop Novelty Reviewer, a literature-aware framework that learns reviewer-aligned evaluation behavior and grounds its judgments in concept-level comparisons to prior work. Our experiments show that the resulting model produces calibrated, interpretable assessments, outperforms strong baselines, detects conceptual plagiarism, and generalizes to an unseen venue. We release the dataset, model, and code to support further research toward more consistent, transparent, and scalable peer review.

\begin{credits}
\subsubsection{\ackname}{This work was partially supported by the Federal Ministry of Research, Technology and Space of Germany [project name: EMuLE – Enhancing Data and Model Efficiency in Multimodal Learning; grant number 16IS24059] and the Leibniz Young Investigator Grant program (project ARENA, LYIG-2023-01) of Leibniz University Hannover, funded by the Ministry of Science and Culture of Lower Saxony (MWK) (grant no. 11-76251-114/2022). The last author was partially funded by the Lower Saxony Ministry of Science and Culture (MWK) with funds from the Volkswagen Foundation’s Zukunft Niedersachsen program [project name: CAIMed - Lower Saxony Center for Artificial Intelligence and Causal Methods in Medicine; grant number: ZN4257]. The authors acknowledge the Hannover Medical School for providing MHH-HPC resources that have contributed to the results reported in this paper.}
\subsubsection{\discintname}{The authors have no competing interests to declare that are relevant to the content of this article.}
\end{credits}
%
% ---- Bibliography ----
%
% BibTeX users should specify bibliography style 'splncs04'.
% References will then be sorted and formatted in the correct style.
%
 \bibliographystyle{splncs04}
\bibliography{custom}

\end{document}